%% file: root.tex
\documentclass[letterpaper, 10 pt, journal, twoside]{IEEEtran}

\ifCLASSINFOpdf
  \usepackage[pdftex]{graphicx}
\else
  \usepackage[dvips]{graphicx}
\fi
\usepackage{amsmath}
\usepackage{amsfonts}
\input{preamble}

\newcommand\copyrighttext{\footnotesize \textcopyright~2025 IEEE.  Personal use of this material is permitted.  Permission from IEEE must be obtained for all other uses, in any current or future media, including reprinting/republishing this material for advertising or promotional purposes, creating new collective works, for resale or redistribution to servers or lists, or reuse of any copyrighted component of this work in other works.
}%

\newcommand\copyrightnotice{%
	\begin{tikzpicture}[remember picture,overlay]
	\node[anchor=south,xshift=0pt,yshift=2pt] at (current page.south) {\fbox{\parbox{\dimexpr\textwidth-\fboxsep-\fboxrule\relax}{\copyrighttext}}};
	\end{tikzpicture}%
}

\begin{document}
\title{\methodName{}: A Flexible Generative Perception Error Model \\ for Probing Self-Driving Planners}
\author{Niklas Hanselmann$^{1,2,3}$, Simon Doll$^{1,2}$, Marius Cordts$^{1}$, Hendrik P.A. Lensch$^{2}$ and Andreas Geiger$^{2,3}$%
\thanks{Manuscript received: November, 3rd, 2024; Revised March, 7th, 2025; Accepted April, 2nd, 2025.}
\thanks{This paper was recommended for publication by Editor Markus Vincze upon evaluation of the Associate Editor and Reviewers' comments.
Niklas Hanselmann and Simon Doll were supported by the German Federal Ministry for Economic Affairs and Climate Action (KI Delta Learning: project number 19A19013A). Andreas Geiger was supported by the ERC Starting Grant LEGO-3D (850533) and the DFG EXC number 2064/1 - project number 390727645.} 
\thanks{$^{1}$ Mercedes-Benz AG R\&D, Sindelfingen, Germany
        {\tt\footnotesize first.last@mercedes-benz.com}}%
\thanks{$^{2}$ University of Tübingen, Tübingen, Germany}%
\thanks{$^{3}$ Tübingen AI Center, Tübingen, Germany}%
\thanks{Digital Object Identifier (DOI): 10.1109/LRA.2025.3562789.}
}

\markboth{IEEE Robotics and Automation Letters. Preprint Version. Accepted April, 2025}
{Hanselmann \MakeLowercase{\textit{et al.}}: \methodName{}} 

\maketitle

\input{sections/0_abstract}  


\copyrightnotice

\input{sections/1_intro}
\input{sections/2_related}
\input{sections/3_method}
\input{sections/4_experiments}
\input{sections/5_conclusion}

\IEEEpeerreviewmaketitle

\ifCLASSOPTIONcaptionsoff
  \newpage
\fi

\bibliographystyle{IEEEtran}
\bibliography{bibliography_short, bibliography, bibliography_custom}

\end{document}

%% file: preamble.tex
\usepackage[dvipsnames]{xcolor}
\usepackage[nolist, nohyperlinks]{acronym}
\usepackage{siunitx}

\usepackage{colortbl}
\usepackage{booktabs}
\usepackage{cleveref}
\usepackage{pifont}
\usepackage[caption=false]{subfig}
\usepackage{tikz}
\input{shortcuts}
\input{acronyms}

%% file: shortcuts.tex
\newcommand{\bb}{\mathbf{b}}\newcommand{\bB}{\mathbf{B}}

\newcommand{\bq}{\mathbf{q}}\newcommand{\bQ}{\mathbf{Q}}

\newcommand{\bs}{\mathbf{s}}\newcommand{\bS}{\mathbf{S}}

\newcommand{\bz}{\mathbf{z}}\newcommand{\bZ}{\mathbf{Z}}

\newcommand{\btau}{\boldsymbol{\tau}}

\newcommand{\nE}{\mathbb{E}}

\newcommand{\nR}{\mathbb{R}}

\newcommand{\cC}{\mathcal{C}}

\newcommand{\cL}{\mathcal{L}}
\newcommand{\cM}{\mathcal{M}}
\newcommand{\cN}{\mathcal{N}}

\newcommand{\cR}{\mathcal{R}}

\newcommand{\figref}[1]{Fig.~\ref{#1}}
\newcommand{\secref}[1]{Section~\ref{#1}}

\newcommand{\eqnref}[1]{Eq.~\eqref{#1}}
\newcommand{\tabref}[1]{Table~\ref{#1}}

\DeclareMathOperator*{\argmax}{argmax~}

\newcommand{\boldparagraph}[1]{\vspace{0.2cm}\noindent{\bf #1:} }

\definecolor{darkgreen}{rgb}{0,0.7,0}

\newcommand{\methodName}{\textsc{Emperror}}

\definecolor{ellisred}{rgb}{0.87,0.44,0.38} 
\definecolor{ellisgreen}{rgb}{0.69,0.90,0.52} 
\definecolor{elliscyan}{rgb}{0.29,0.77,0.74} 
\definecolor{ellisorange}{rgb}{0.89,0.55,0.28} 
\definecolor{ellisblue}{rgb}{0.41,0.61,0.86} 

\definecolor{gtred}{rgb}{0.639, 0.078, 0.078}
\definecolor{predblue}{rgb}{0, 0.266, 0.435}

\definecolor{Gray}{gray}{0.9}
\newcommand{\highlight}{\cellcolor{Gray}}

%% file: acronyms.tex
\begin{acronym}
    \acro{il}[IL]{imitation learning}
\end{acronym}

\begin{acronym}
    \acro{bev}[BEV]{bird's-eye view}
\end{acronym}

\begin{acronym}
    \acro{cvae}[CVAE]{conditional variational autoencoder}
\end{acronym}

\begin{acronym}
    \acro{pem}[PEM]{perception error model}
\end{acronym}

\begin{acronym}
    \acro{elbo}[ELBO]{evidence lower bound}
\end{acronym}

\begin{acronym}
    \acro{cnn}[CNN]{convolutional neural network}
\end{acronym}

\begin{acronym}
    \acro{mlp}[MLP]{multilayer perceptron}
\end{acronym}

\begin{acronym}
    \acro{sota}[SOTA]{\textit{state-of-the-art}}
\end{acronym}

\begin{acronym}
    \acro{mate}[mATE]{mean Average Translation Error}
    \acro{maoe}[mAOE]{mean Average Orientation Error}
    \acro{mase}[mASE]{mean Average Scale Error}
    \acro{mave}[mAVE]{mean Average Velocity Error}
    \acro{ap}[AP]{Average Precision}
    \acro{cd}[CD]{Cumulative Absolute Difference Area}
    \acro{cr}[CR]{Collision Rate}
    \acro{ade}[ADE]{Average Displacement Error}
    \acro{fde}[FDE]{Final Discplacement Error}
\end{acronym}

%% file: sections/0_abstract.tex
\begin{abstract}
To handle the complexities of real-world traffic, learning planners for self-driving from data is a promising direction. 
While recent approaches have shown great progress, they typically assume a setting in which the ground-truth world state is available as input. 
However, when deployed, planning needs to be robust to the long-tail of errors incurred by a noisy perception system, which is often neglected in evaluation.
To address this, previous work has proposed drawing adversarial samples from a \ac{pem} mimicking the noise characteristics of a target object detector. However, these methods use simple \acp{pem} that fail to accurately capture all failure modes of detection. 
In this paper, we present \methodName{}, a novel transformer-based generative \ac{pem}, apply it to stress-test an \ac{il}-based planner and show that it imitates modern detectors more faithfully than previous work. Furthermore, it is able to produce realistic noisy inputs that increase the planner's collision rate by up to \SI[detect-weight]{85}{\percent}, demonstrating its utility as a valuable tool for a more complete evaluation of self-driving planners. 
\end{abstract}

%% file: sections/1_intro.tex
\section{Introduction}
\IEEEPARstart{A}{fter} years of progress, autonomous driving systems are able to handle increasingly complex situations~\cite{Janai2020}. This is enabled, in part, by solving several aspects of driving with learned modules, such 
as perception~\cite{Wang2021CORL, Wang2023ICCV, Li2023Toponet}
and motion forecasting~\cite{Salzmann2020ECCV, Hu2021FIERY}.
Recently, there has been increased interest in managing the complexity of human behavior in traffic by also learning the planning task~\cite{Renz2022CORL, Dauner2023Parting}, which has been accelerated through the emergence of motion forecasting- and 
planning-centric benchmarks and datasets~\cite{Caesar2021CVPRWORK, Montali2023ARXIV}.
Most of this work assumes a simplified setting where the ground-truth world state is available as input and focuses on accuracy of the planned trajectory with respect to human driving demonstrations. 
As a result, robustness to the residual risk of errors in the perception system, which is ultimately just an imperfect model operating on incomplete observations of the world, remains underexplored.

\begin{figure}[t]
    \centering
    \includegraphics[width=\columnwidth, trim={1.5cm 0.3cm 1.5cm 0.2cm},clip]{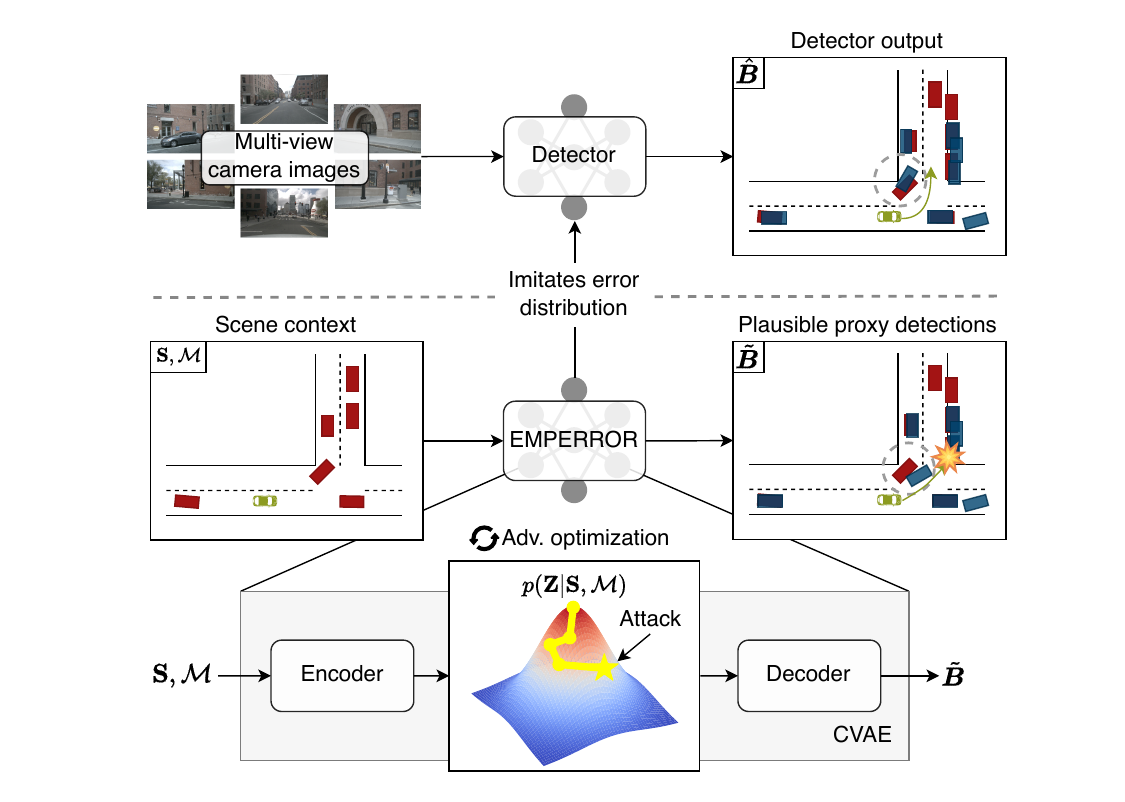}
    \caption{\textbf{Method Overview.} We propose \methodName{}, a generative model that imitates a given detector by modeling the distribution of its perception errors conditioned on a ground-truth state and BEV map as scene context. Adversarial optimization in the model's latent space can then produce challenging yet plausible proxy detections from that distribution which stress-test the robustness of a given planner, e.g. by inducing collisions.}
    \label{fig:teaser}
\end{figure}

In this work, we aim to illuminate the susceptibility of learned planning, which is often brittle in the face of o.o.d inputs~\cite{Filos2020ICML}, to these errors. Recent seminal studies~\cite{Innes2023ICRA, Sadeghi2023ARXIV} have approached this problem from the lens of adversarial attacks. 
To this end, they first construct a \acf{pem}~\cite{Piazzoni2021IJCAI, Sadeghi2021NEURIPSW}, which allows sampling multiple noisy estimates imitating a target 3D object detector given a ground-truth scene representation as context. 
Then, by leveraging the \ac{pem} as a proxy of the detector, challenging samples that stress-test the target planner can be found by employing an adversarial search strategy. 
While promising, these works consider simple, synthetic scenes that do not capture the complexities of real-world data. 
Moreover, they employ simple \acp{pem} that phrase the error modeling task as an isolated, per-object perturbation of the ground-truth state, rather than jointly reasoning over the entire scene context. 
Hence, they cannot faithfully model the intricacies of the error patterns exhibited by modern 3D object detectors, such as duplicate detections resulting in false-positives and correlations in errors for groups of objects.

Motivated by this, we propose \methodName{}, a novel generative \textbf{emp}irical \textbf{error} model based on the transformer architecture, that can more faithfully capture the error characteristics of a target detector. 
Our key idea is to leverage the attention mechanism and a flexible set of latent queries to model the full range of failure modes, including false-positives, in a scene-consistent manner. 
Furthermore, \methodName{} provides a prior over different error patterns for a given scene context, enabling us to draw adversarial, yet plausible samples to stress-test a target planner. 
Building on these advantages, we design a framework to probe the robustness of learned planners to noisy perception inputs, which is visualized in \figref{fig:teaser}. 
We then apply the proposed framework to an \acf{il}-based planner, modeling three different modern camera-based 3D object detectors, and show learned planning is indeed vulnerable to plausible noise from the long-tail of perception errors. 
We believe \methodName{} can serve as a valuable tool for data-driven evaluation of self-driving planners.

\boldparagraph{Contributions}
(1)~We propose \methodName{}, a novel flexible, transformer-based, generative \ac{pem} for probing planning, that can more faithfully imitate modern object detectors than previous work. 
(2)~We propose and integrate \methodName{} into a framework for probing the robustness of a planner to noise in its perception system. 
(3)~We demonstrate that this framework can reveal unsafe behavior in learned planning, even for minor, plausible detection errors.

%% file: sections/2_related.tex
\section{Related Work}
\boldparagraph{Perception Error Models}
Accurately modeling the noise characteristics of a perception module enables an analysis of typical failure modes and informs the design of downstream modules.
Several works \cite{Mitra2018ITSC,  Piazzoni2021IJCAI, Piazzoni2023ARXIV}
rely on classical statistical models often coupled with elements of manual design to capture regression errors and false-negatives in detection tasks. 
While these approaches yield lightweight, easily interpretable \acp{pem}, they are limited in the fidelity and complexity of noise patterns that can be modeled. Recent work addresses this limitation by instantiating the \ac{pem} through a neural network. In~\cite{Innes2023ICRA}, the authors use a feed-forward network to model false-negative detections.
In \cite{Sadeghi2021NEURIPSW, Sadeghi2023ARXIV}, the authors model spatial errors and false-negatives using a probabilistic feed-forward network to enable efficient evaluation of decision making modules in simulation. 
In \cite{Wong2020ECCV, Ju2023PerceptionImitation}, the authors propose fully-convolutional \acp{pem} that mimic the error characteristics of 
LiDAR-based detectors \cite{Yang2018PIXOR, Yin2021Centerpoint}.
As they resemble the overall architecture of a standard convolutional one-stage detector, they can model regression errors, false-negatives and false-positives but are not probabilistic and thus do not permit sampling.
In summary, the \acp{pem} proposed in previous work either do not fully capture the error characteristics of a target perception system, do not permit sampling, or both. As these are prerequisites to find plausible worst-case perception errors for planning, we propose a novel \ac{pem} that satisfies these requirements.

\boldparagraph{Generating Safety-Critical Scenarios}
When deployed, self-driving systems are required to robustly handle rare and potentially safety-critical scenarios from the long-tail of driving. 
Since real-world data collection is limited in scalability and diversity, there has been increased interest in the automated generation of safety-critical scenarios in recent years. 
The majority of this work focuses on automatically altering the behavior of other traffic participants to induce failure in the target autonomy system \cite{Ding2020IROS, Suo2023CVPR, Wang2021CVPRb, Rempe2022CVPR, Hanselmann2022ECCV}. 
Rather than testing against external long-tail behavior, we instead look inward to examine the effects of long-tail noise in the autonomy system's own perception system by sampling from a \ac{pem}.
While this high-level idea has been explored before, previous work~\cite{Innes2023ICRA, Sadeghi2023ARXIV} uses simple \acp{pem} that do not fully capture all dependencies and failure modes, and tests in simple, synthetic scenes that do not reflect the complexity of real-world data.
In contrast, we propose a novel transformer-based \ac{pem} that fully models the false-positive, false-negative and regression error characteristics of a given 3D-object detector and apply our framework on challenging real-world data.

%% file: sections/3_method.tex
\section{Method}
\boldparagraph{Problem Statement}
We are interested in supplementing offline evaluation by stress-testing modular autonomy systems, which we will consider to consist of a 3D object detector as the perception module and a downstream planner in this study.
Let us denote the true world state as a set of vectors $\bS = \{\bs_0, \cdots, \bs_N\} \in \nR^{d_s}$ describing the position, heading angle, spatial dimensions, first- and second-order longitudinal- and angular dynamics in the ego vehicle's frame as well as the semantic class for each of $N$ objects in the scene. 
The object detector then processes raw sensor observations of the world into a set of 3D bounding boxes \mbox{$\hat{\bB} = \{\hat{\bb}_0, \cdots, \hat{\bb}_{\hat{N}}\} \in \nR^{d_b}$} similar to $\bS$, but with the dynamics reduced to a velocity vector. 
Based on the 3D bounding boxes, as well as a rasterized \ac{bev} map $\cM \in \cR^{h \times w \times 5}$ containing information on the static scene layout, the planner then computes a future trajectory $\btau$ to be driven.

The detector incurs errors in the form of false-positives, false-negatives and inaccurate regression of bounding box parameters when compared to the ground-truth set derived directly from $\bS$. 
These errors are individual to the specific model architecture and weights in question, and often depend on the configuration of the scene, e.g. due to correlations with the relative object pose, correlations among groups of objects or correlations between objects and map elements.
To probe the planner's sensitivity to this imperfect perception, we would like to find worst-case, but plausible errors for a given scene that drive the planned trajectory towards violation of safety constraints. 
To this end, we learn a conditional generative \ac{pem} as a proxy that imitates the detector given the ground-truth scene context $(\bS, \cM)$ as input. By repeatedly drawing samples $\tilde{\bB} = \{\tilde{\bb}_0, \cdots, \tilde{\bb}_{\tilde{N}}\} \in \nR^{d_b}$ from this proxy and measuring the quality of the corresponding planned trajectory, we aim to gauge the influence of noise patterns and optimize for planning failures, such as a collision.

\begin{figure*}[t!]
    \centering
    \includegraphics[width=\linewidth, trim={1cm 0.4cm 0 0cm},clip]{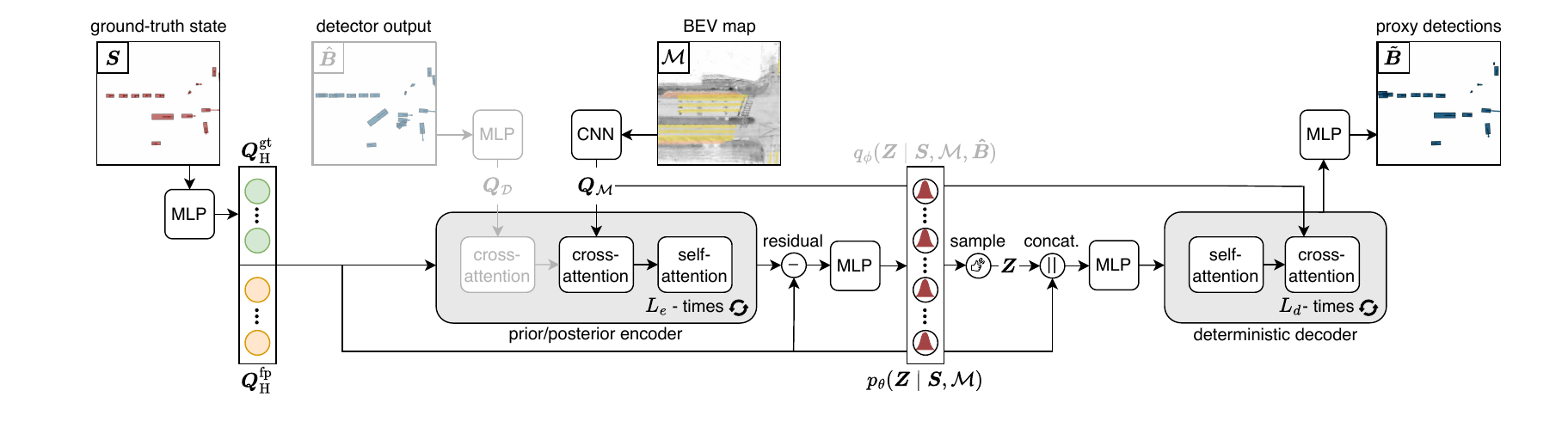}
    \caption{\textbf{Generative Perception Error Model (PEM).} We propose a \ac{pem} based on the \acf{cvae} framework to model the error characteristics of a target detector. It consists of a \textit{prior encoder}, inferring a distribution over the latent variable $\bZ$ given a ground-truth state $\bS$ and BEV map $\cM$ as scene context at test-time, and a \textit{deterministic decoder}, which produces a set of proxy detections $\tilde{\bB}$ given $\bZ$. At training-time, a \textit{posterior encoder} with a similar architecture as the prior encoder is used to encode and sample from the latent distribution. In contrast to the prior encoder, it also has access to privileged information in the form of the detector outputs $\hat{\bB}$ to be reconstructed. Privileged components are shown in \textcolor{lightgray}{faint coloring}.}
    \label{fig:architecture}
\end{figure*}

\subsection{Conditional Generative \ac{pem}}
\label{subsec:condgenpem}
We cast the imitation of perception errors given a ground-truth scene state as a set-to-set modeling problem, for which the transformer architecture is a natural choice due to its permutation-invariant modeling of relationships among set elements. We propose a conditional generative model building on a transformer encoder-decoder architecture~\cite{Carion2020ECCV} under the framework of \acp{cvae}~\cite{Sohn2015NIPS}. Specifically, we aim to construct a latent variable model capturing
\begin{equation}
    \label{eq:lvm}
    P\left(\hat{\bB} \mid \bS, \cM\right) = \int P\left(\hat{\bB} \mid \bS, \cM, \bZ\right) P\left(\bZ \mid \bS, \cM\right)  \,d\bZ
\end{equation}
where $\bZ = \{\bz_0, \cdots, \bz_{\hat{N}}\} \in \nR^{d_z}$ is a set of per-object latent variables of dimensionality $d_z$ explaining the stochasticity of the target detector's perception errors by capturing different noise patterns. 
This allows sampling multiple plausible sets of detections $\tilde{\bB} \sim p_\theta(\hat{\bB} \mid \bS, \cM)$ by applying different latent error characteristics sampled from a learned conditional Gaussian prior modeling $\bZ \sim p_\theta (\bZ \mid \bS, \cM)$ to a given scene context $(\bS, \cM)$. This is done via a deterministic decoder $\hat{\bB} = f_\theta(\bZ, \bS, \cM)$.
As evaluating the integral in \eqref{eq:lvm} is intractable, learning under the \ac{cvae} framework utilizes a learned approximation to the posterior $q_\phi (\bZ \mid \bS, \cM, \hat{\bB})$ to maximize the 
\ac{elbo}~\cite{Kingma2014ICLR, Sohn2015NIPS}:
\begin{align}
    \log P\left(\hat{\bB} \mid \bS, \cM \right) \geq \,
    \nE_{q_\phi \left(\bZ \mid \bS, \cM, \hat{\bB} \right)} 
        \left[ \log p_\theta \left( \hat{\bB} \mid \bS, \cM, \bZ \right) \right] \nonumber \\
    - D_\text{KL} \left(
        q_\phi \left(\bZ \mid \bS, \cM, \hat{\bB} \right) 
        \mid\mid  
        p_\theta \left(\bZ \mid \bS, \cM \right) \right)
    \label{eq:elbo}
\end{align}
Although both the prior and the posterior distribution are factorized over objects, the distribution over each latent variable $\bz_n \in \bZ$ is computed from the full set of ground-truth states $\bS$ (and detector outputs $\hat{\bB}$ for the posterior). 
Similarly, the decoder considers the joint sets of per-object latent variables $\bZ$ and states $\bS$ to output individual detections $\tilde{\bb}_n \in \tilde{\bB}$. 
This enables our model to generate realistic, scene-consistent noise patterns by allowing it to capture higher-order relationships. 
In the following, we briefly describe the main components in our architecture, as depicted in \figref{fig:architecture}.

\boldparagraph{Initial Hypotheses}
Both the probabilistic encoders modeling $p_\theta$ and $q_\phi$ as well as the deterministic decoder $f_\theta$ are transformers operating on a shared set of queries \mbox{$\bQ_H = \{\bq_H^0, \cdots, \bq_H^{N_q}\} \in \nR^{d_h}$} representing detection hypotheses.
Specifically, to model true-positives and false-negatives, we initialize a subset $\bQ_H^{gt}$ of $N_q^{gt}$ queries by projecting the ground-truth state of the scene to the feature dimensionality $d_h$ of the model via a \ac{mlp}, which implies that $N_q^{gt}$ is variable depending on the number of objects in the scene.
Under the assumption that for any reasonable detector the majority of ground-truth objects will have a corresponding detector output, this provides the model with a sensible first initialization and simplifies its task to the estimation of a residual to the ground-truth regression parameters and semantic class scores.
To model false-positives, a second subset $\bQ_H^{fp}$ comprised of a fixed number of $N_q^{fp}$ constant learnable embeddings of dimensionality $d_h$ capturing dataset-level statistics is used as these can not be trivially initialized from the scene context. 

\boldparagraph{Prior Encoder}
The prior encoder refines the initial hypotheses $\bQ_H$ while considering the context of the scene layout provided by the map $\cM$ through a series of $L_e$ transformer layers to estimate the parameters of the prior distribution 
\mbox{$p_\theta (\bZ \mid \bS, \cM) = 
    \cN(
        \boldsymbol{\mu}^p_\theta (\bS, \cM),
        \boldsymbol{\sigma}^p_\theta (\bS, \cM)
    )
$}.
The map is represented by a set of feature vectors \mbox{$\bQ_{\cM} = \{\bq_{\cM}^0, \cdots, \bq_{\cM}^{h \times w}\} \in \nR^{d_h}$} obtained as the cells of the feature grid of a convolutional map encoder, to which we add sinusoidal positional embeddings to retain spatial information. 
Through repeated blocks of self-attention among the detection hypotheses and cross-attention to the map features, the initial hypotheses are iteratively adjusted towards representing noisy detections for the current scene context. 
This allows the model to capture crucial relationships, such as occlusion between objects or duplicate detection hypothesis for the same ground-truth object, for example of different semantic classes or at various depths along the viewing ray, a characteristic pattern in camera-based 3D detectors. 
The residual between the initial and refined hypotheses, which captures possible error patterns, is then input to an MLP estimating the mean vectors and diagonal covariance matrices of the factorized prior distribution.

\boldparagraph{Posterior Encoder} 
The posterior encoder estimates the parameters of the approximate posterior $q_\phi (\bZ \mid \bS, \cM,\hat{\bB}) = \cN(\boldsymbol{\mu}^q_\phi (\bS, \cM, \hat{\bB}), \boldsymbol{\sigma}^q_\phi (\bS, \cM, \hat{\bB}))$, which is analogous to the prior, and largely follows the same architecture. 
However, we incorporate the detector outputs $\hat{\bB}$ via an additional cross-attention block in each of its $L_e$ transformer layers. 
To this end, we project them to the model's feature space similar to $\bQ_H^{gt}$ via a separate \ac{mlp} to obtain a set of detection features $\bQ_D = \{\bq_D^0, \cdots, \bq_D^{N_D}\} \in \nR^{d_h}$.

\boldparagraph{Decoder}
Given the set of initial hypotheses and map context, the decoder applies a sampled latent error pattern $\bZ$ to produce a set of proxy bounding boxes $\tilde{\bB} = f_\theta(\bZ, \bS, \cM)$. 
To this end, the information in both $\bQ_H$ and $\bZ$ is first fused by applying an \ac{mlp} to the feature-wise concatenation of each pair of vectors $(\bq_n, \bz_n)$, forming $\tilde{\bQ}_H \in \cR^{d_h}$.
Similarly to the prior encoder, $\tilde{\bQ}_H$ is then refined through a series of $L_d$ transformer layers performing self-attention and cross-attending to $\bQ_{\cM}$. Finally, the refined latent queries are decoded to the bounding box regression parameters and independent per-class sigmoid classification scores. This mirrors the prevalent output configuration used in state-of-the-art 3D 
detectors~\cite{Wang2021CORL, Li2022ECCV, Wang2023ICCV}.

\boldparagraph{Training}
We optimize the model on a dataset of tuples $(\bS, \cM, \hat{\bB})$ collected by running inference with a target detector on the corresponding sensor data to obtain noisy detections.
Similar to prior work~\cite{Suo2021CVPR, Rempe2022CVPR},
we use a modified CVAE objective $\cL_{\text{cvae}} = \cL_{\text{recon}} + \beta \cL_\text{JS}$ consisting of a reconstruction error and a divergence regularizer:
\begin{align}
        \cL_{\text{recon}} = & \sum_{(i,j) \in \Omega}\left(\lVert\tilde{\bb}_i^{reg} - \hat{\bb}_j^{reg}\rVert_1 + \sum_{c=0}^{N_c} \text{BCE}\left(\tilde{\bb}_{i,c}^{cls}, \hat{\bb}_{j,c}^{cls}\right) \right) \nonumber \\ 
            &-  \sum_{i \in \emptyset} \sum_{c=0}^{N_c} \log (1 - \tilde{\bb}_{i,c}^{cls})\label{eq:recon_loss}\\
        \cL_{\text{JS}} = & \text{JS}^{G_\alpha} \left(
            q_\phi \left(\bZ \mid \bS, \cM, \hat{\bB}\right) 
            \mid\mid
            p_\theta \left(\bZ \mid \bS, \cM\right)
        \right)  \ .\label{eq:divergence_reg}
\end{align}
For the reconstruction loss, we first compute correspondences between the set of boxes $\tilde{\bB}$ drawn from the \ac{pem} and the set of boxes $\hat{\bB}$ produced by the target detector. 
The correspondences for the subset of boxes produced from ground-truth initialized hypotheses $\bQ_H^{gt}$ are obtained by greedily matching detections $\hat{\bb_n}$ to ground-truth bounding boxes $\bb_n$ via the logic proposed in~\cite{Caesar2019ARXIV} and kept fixed throughout training. 
For those left unmatched, which includes the boxes produced from $\bQ_H^{fp}$, we obtain an assignment of the remaining boxes in $\hat{\bB}$ that have not been matched in the previous step online via the Hungarian algorithm, following the standard practice in~\cite{Wang2021CORL, Li2022ECCV, Wang2023ICCV}. 
This set of correspondences between predicted hypothesis and detector targets is termed $\Omega$.
Any boxes in $\tilde{\bB}$ left without correspondence after these steps are treated as belonging to the no-object set $\emptyset$.

The reconstruction loss is then computed as the sum of an $l_1$-term for the regression parameters and the binary cross-entropy for the per-class sigmoid scores. Unmatched proxy bounding boxes are expected to have a classification score of zero.
To discourage the prior encoder from placing probability mass in regions assigned low likelihood by the privileged approximate posterior, we replace the forward KL-divergence in \eqref{eq:elbo} with the skew-geometric Jensen-Shannon divergence $\text{JS}^{G_\alpha}$~\cite{Deasy2020NEURIPS}. Intuitively, this allows us to interpolate between the forward KL, which encourages \textit{mode-covering} behavior, and the backward KL, which encourages \textit{mode-seeking} behavior, via a parameter $\alpha$. 
This formulation permits trading off diversity for a decreased chance of sampling latent error patterns from the prior that would be unlikely under the posterior at test time. 
Finally, we use a weighting factor $\beta$ to control the strength of the divergence regularizer in the overall objective, as proposed in~\cite{Higgins2017ICLR}.

\subsection{Imitation Learning-based Planner}
\label{subsec:planner}
We consider a simple transformer-based planner that is prototypical of the planning modules proposed in recent literature~\cite{Renz2022CORL, Hu2023CVPR, Jiang2023ICCV, Doll2024DualAD}. 
It operates on \ac{bev} projections of $\hat{\bB}$ or $\tilde{\bB}$, which are encoded via an \ac{mlp} to form $\bQ_H^\pi \in \nR^{d_\pi}$, as well as a set of map features $\bQ_{\cM}^\pi \in \nR^{d_\pi}$ obtained from a convolutional encoder.
The planner then forms a latent plan by refining an initial constant learnable embedding $\bq_\text{ego}^\pi$ through a series of $L_\pi$ transformer layers consisting of two cross-attention blocks attending to the encoded detection results $\bQ_H^\pi$ and map features $\bQ_{\cM}^\pi$. 
To provide additional context, we add embeddings of the current speed of the ego vehicle $\vartheta^\text{ego}$, as well as a high-level navigation command
$c^\text{nav}$ 
(i.e. \textit{go-straight}, \textit{turn-left} or \textit{turn-right}) to the initial $\bq_\text{ego}^\pi$.
Finally, an \ac{mlp} decodes a trajectory $\hat{\btau} = \pi_\omega(\bB, \cM, \vartheta^\text{ego}, c^\text{nav}) \in \nR^{T_\text{plan} \times 3}$ of future waypoints consisting of a \ac{bev} position and heading angle. The model is trained via standard imitation-learning on expert trajectories using an $l_1$-loss.

\subsection{Probing Planning}
\label{subsec:probing_planning}
Given a scene- $(\bS, \cM)$ and planning context $(\vartheta^\text{ego}, c^\text{nav})$, we now aim to probe $\pi_\omega$ with respect to its sensitivity to noise in its input perception results. 
To this end, we leverage our generative \ac{pem} in an adversarial fashion to draw samples $\tilde{\bB} \sim p_\theta(\hat{\bB} \mid \bS, \cM)$ that induce failure in $\pi_\omega$. 
Specifically, for a driving cost $\cC$ measuring the quality of the generated trajectory, we formulate this process as an optimization problem:
\begin{align}
\begin{split}
    \tilde{\bB}^{\ast} = & \argmax_{\bZ \sim p_\theta (\bZ \mid \bS, \cM)} \mkern-10mu \cC\left(\pi_\omega(f_\theta\left(\bZ, \bS, \cM \right), \cM, \vartheta^\text{ego}, c^\text{nav}), \bS, \cM\right) \\
    & \quad \text{s.t.} \quad - \kappa \, \boldsymbol{\sigma}^p_{\theta,n} \leq \bz_n - \boldsymbol{\mu}^p_{\theta,n} \leq  \kappa \, \boldsymbol{\sigma}^p_{\theta, n} \quad \forall \ \bz_n \in \bZ
    \label{eq:adv_opt}
\end{split}
\end{align}
Here, the latent space of our generative model forms the search space, which is particularly amenable to adversarial optimization due to the explicit prior over $\bZ$: 
By bounding the maximum standard deviation we allow each $\bz_n$ to move from the mean, the solution space can be constrained to high-likelihood regions, ensuring plausibility.

\boldparagraph{Objective} 
For this study, we choose a simple collision cost measuring the closest Euclidean distance $d(\hat{\btau}^t, \bs^t_n)$ between the \ac{bev} center point of any other object and a planned waypoint in $\hat{\btau}$ within the planning horizon $T_\text{plan}$. Furthermore, we add a regularizer encouraging $\bZ$ to remain likely under the prior, resulting in the following overall cost:
\begin{equation}
    \cC = - \min_{\substack{n \in \{0, \cdots, N\} \\ t \in \{0, \cdots, T_\text{plan}\}}} d(\hat{\btau}^t, \bs^t_n) + 
    \lambda \sum_{n=0}^{N_q} \log p_\theta (\bz_n \mid \bS, \cM)
    \label{eq:adv_cost}
\end{equation}
where $\lambda$ controls the strength of the prior regularization.

\boldparagraph{Optimization}
Since all components in our framework are differentiable, we approach the optimization problem in \eqref{eq:adv_opt} via gradient ascent. This results in the following procedure:

\begin{enumerate}
    \item Infer the prior distribution $p_\theta (\bz_n \mid \bS, \cM)$ for the current scene via the corresponding probabilistic encoder, and initialize all $\bz_n$ to the mean $\boldsymbol{\mu}^p_{\theta, n} (\bS, \cM)$.
    \item Take $N_\text{opt}$ gradient steps on $\bZ$ maximizing the cost function in \eqref{eq:adv_cost} and clamp all $\bz_n$ to the permitted interval $(-\kappa \, \boldsymbol{\sigma}^p_{\theta, n}, \kappa \, \boldsymbol{\sigma}^p_{\theta, n})$ after each update to satisfy the constraint in \eqref{eq:adv_opt}.
    \item After each step, determine if the planned trajectory results in a future collision with any object via an intersection check on the \ac{bev} bounding-boxes of all ego-object pairs. If no collision is found within $N_\text{opt}$ iterations, the optimization terminates unsuccessfully.
\end{enumerate}

To generate multiple challenging perception results per scene, we run the above procedure up to $N_\text{trial}$ times, each time removing the object used to compute the collision cost in \eqref{eq:adv_cost} in the previous trial from consideration.

%% file: sections/4_experiments.tex
\section{Experiment Results}
In this section, we experimentally analyze \methodName{} in terms of (1) its effectiveness in faithfully imitating modern camera-based 3D object detectors and (2) illuminate its utility in probing the robustness of planning to such errors.

\begin{table}
\centering
\caption{\textbf{Performance evaluation for different \acp{pem}}. The comparison of \ac{pem} and detector characteristics is measured in terms of \acf{cd} for different perception metrics. $\ast$ denotes a variant without visibility input, while $\dagger$ denotes additional map input.}
\resizebox{\columnwidth}{!}{
\begin{tabular}{lcccc}
    \toprule
    Model &   \highlight{\ac{cd}-mPrec.} &  \ac{cd}-mATE &  \ac{cd}-mAOE &  \ac{cd}-mAVE \\
    \midrule
    \textbf{Detr3d}~\cite{Wang2021CORL} & \highlight{ }\\
    Static Gauss       & \highlight{0.082}  &     0.479 &      0.453 &    0.601 \\
    MLP + Gauss $\ast$ & \highlight{0.057} &      0.116 &      0.059 &    0.310 \\
    MLP + Gauss & \highlight{0.061} &      0.120 &       0.064 &    0.204 \\
    MLP + Gauss $\dagger$ & \highlight{0.065} &      0.162 &       0.102 &    0.263 \\
    ResNet + Gauss & \highlight{0.062} &      0.138 &       0.068 &    0.274 \\
    ResNet + StudT & \highlight{0.052} &      0.117 &       0.072 &    0.333 \\
    ResNet + StudT $\dagger$ & \highlight{0.069} &      0.156 &       0.078 &    0.351 \\
    \textsc{Ours} & \highlight{\textbf{0.038}} &      \textbf{0.053} &       0.060 &    0.133 \\
    \ \ w/ KL-Divergence & \highlight{0.044} &      0.061 &       0.066 &    0.140 \\
    \ \ w/ $N_q^{fp}=0$ & \highlight{0.051} &      0.115 &       \textbf{0.056} &    0.182 \\
    \ \ w/ $N_q^{fp}=256$ & \highlight{\textbf{0.038}} &      0.055 &       0.058 &    \textbf{0.128} \\
    \midrule
    \textbf{BEVFormer}~\cite{Li2022ECCV} & \highlight{ }\\
    MLP + Gauss & \highlight{0.076} &      0.152 &       0.076 &    0.145 \\
    ResNet + Studt &  \highlight{0.068} &      0.142 &       \textbf{0.061} &    0.466 \\
    \textsc{Ours} & \highlight{\textbf{0.046}} &      \textbf{0.069} &       0.069 &    \textbf{0.044} \\
    \midrule
    \textbf{StreamPETR}~\cite{Wang2023ICCV} & \highlight{ }\\
    MLP + Gauss & \highlight{0.066} &      0.122 &       0.088 &    0.089 \\
    ResNet + Studt & \highlight{0.089} &      0.191 &       0.100 &    0.342 \\
    \textsc{Ours} & \highlight{\textbf{0.056}} &      \textbf{0.109} &       \textbf{0.087} &    \textbf{0.046} \\
\bottomrule
\end{tabular}
}
\label{tab:pem_results}
\end{table}

\boldparagraph{Dataset}
We utilize the challenging and well-established nuScenes dataset~\cite{Caesar2019ARXIV} consisting of 1000 real-world sensor logs covering a diverse range of scenarios, each \SI{20}{\second} in length, and tracked 3D annotations of ten different object categories at a frequency of \SI{2}{\hertz}.
We use the official detection sub-split of the training set to train the detector and apply it on the tracking sub-split to generate training data for the \ac{pem}. 
We evaluate both on the official validation split.
\begin{figure*}
    \centering
    \includegraphics[width=\textwidth]{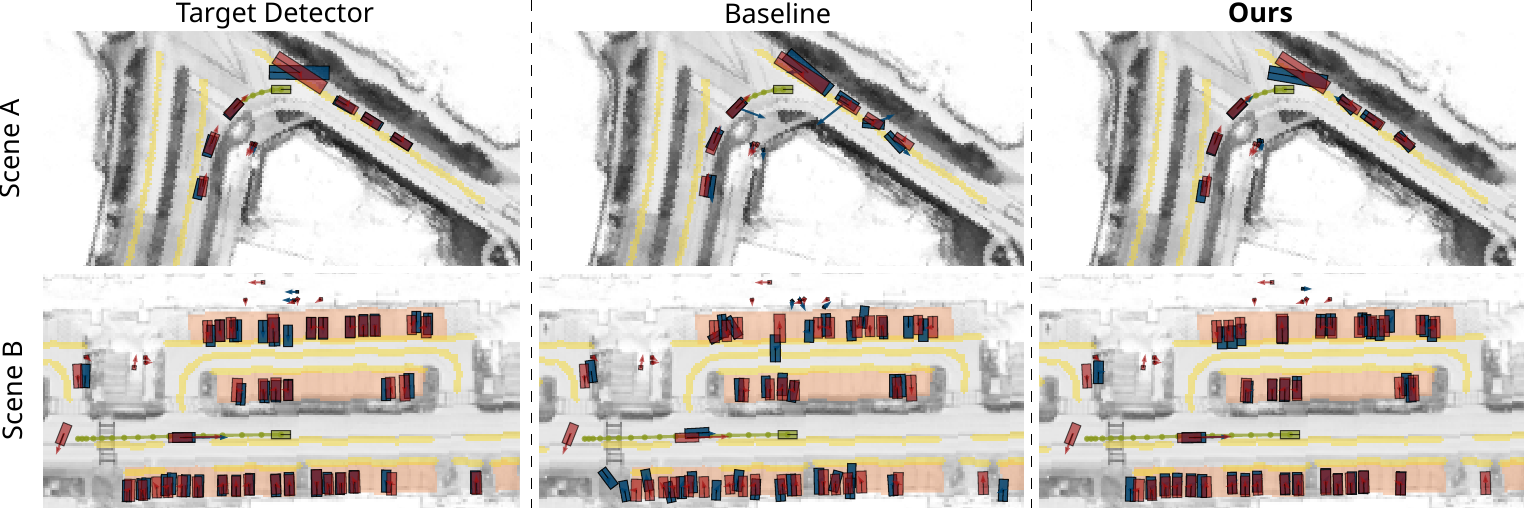}
    \caption{\textbf{\ac{pem} Qualitative Results.} Perception errors modeled by a baseline \ac{pem} (middle) using an MLP with a Gaussian and \methodName{} (right) compared to DETR3D detections (left). \textcolor{gtred}{Red boxes} indicate ground truth objects, \textcolor{predblue}{blue boxes} the model predictions. 
    While the baseline model samples implausible perception velocities and does not adapt to scene context, such as the parking area in Scene B, our approach closely mimics the target detector.}
    \label{fig:quali_pem}
\end{figure*}

\boldparagraph{Target Detectors}
To include a variety of error characteristics in our evaluation, we choose three different modern detectors: (1) DETR3D~\cite{Wang2021CORL} based on sparse object queries, (2) StreamPETR~\cite{Wang2023ICCV} that utilizes temporal object queries that are propagated through time and (3) BEVFormer~\cite{Li2022ECCV} which utilizes an intermediate temporal BEV-feature grid. All detectors have been trained for 48 epochs, utilizing a ResNet-101~\cite{he2016deep} backbone and the official implementations.
\begin{figure}
    \centering
    \subfloat[Car]{\includegraphics[width=0.49\columnwidth]{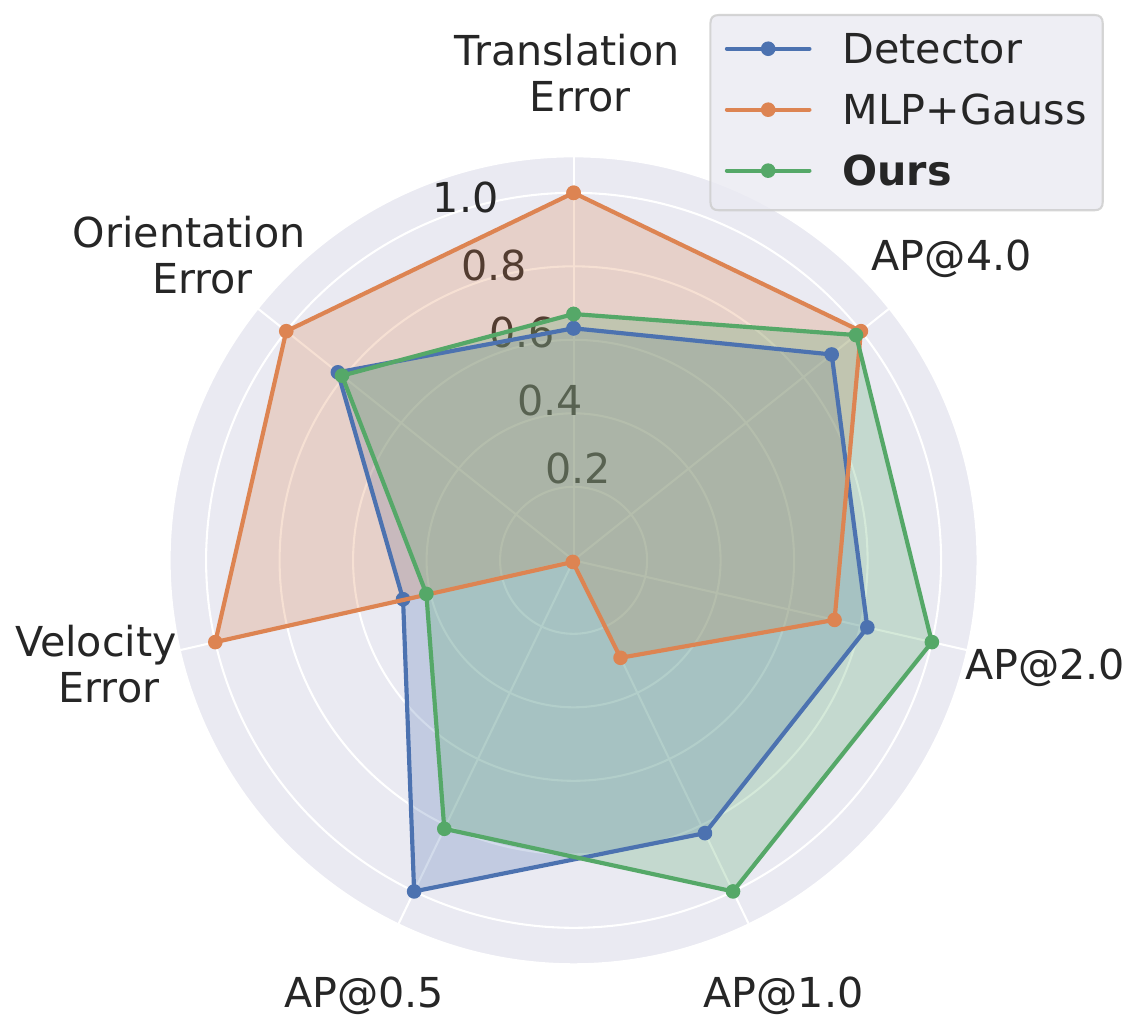}\label{fig:spider_plots:car}}
    \hfill
    \subfloat[Pedestrian]{\includegraphics[width=0.49\columnwidth]{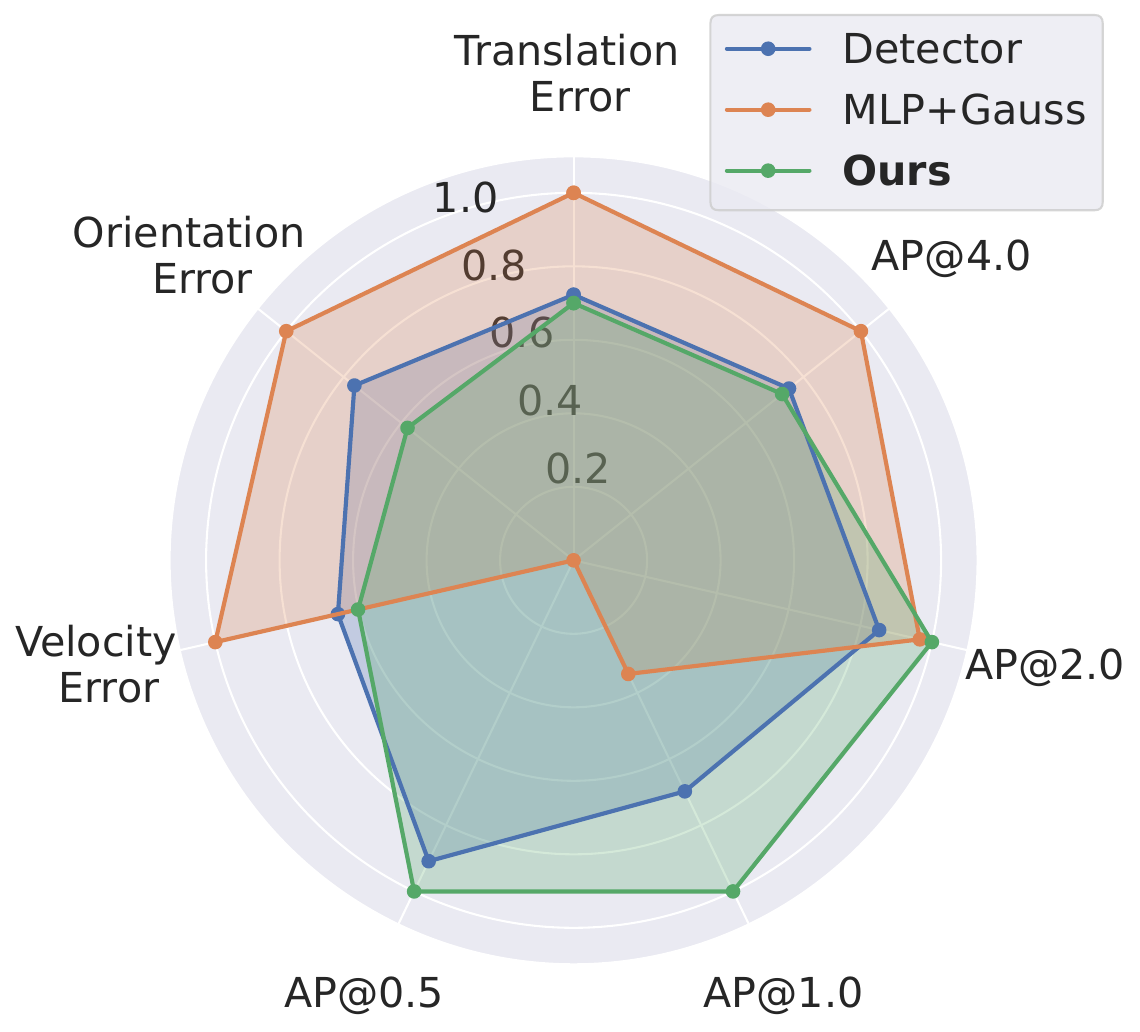}\label{fig:spider_plots:pedestrian}}\\
    \caption{\textbf{\ac{pem} Error Characteristics.} 
    We show the precision and mean regression errors for DETR3D, \methodName{} and the MLP + Gauss configuration as baseline. All metrics are normalized for clearer visual comparison.}
    \label{fig:spider_plots}
\end{figure}

\subsection{Evaluation of Error Imitation Quality}
\label{subsec:pem_eval}

\boldparagraph{Metrics}
To measure the perception performance, we follow the official metric definitions of the nuScenes benchmark. These include the \ac{ap} as well as regression error metrics for true positives. The latter include the \ac{mate}, \ac{maoe} and \ac{mave}. 
For the exact metric definitions, we refer the reader to~\cite{Caesar2019ARXIV}.
However, analyzing the mean precision and error values alone is insufficient, as it does not quantify how the errors evolve with decreasing detection confidence values. To this end, we compare the integral of absolute differences in metric values over all recall intervals between the target detector and \ac{pem}, which we term \acf{cd}.

\boldparagraph{Baselines}
We implement a simple baseline similar to the observation noise used in the Kalman filter, which models the perception error distribution as a scene-independent Gaussian distribution. To this end, we compute the per-class empirical Gaussian over 3D bounding box regression errors and class logits for detections with a ground-truth match on the \ac{pem} sub-split of the training dataset. Samples from this distribution are then applied to ground-truth bounding boxes to create noisy perception results. We additionally compute the per-class false-negative rate, at which we randomly drop detections. Since this baseline, which we term Static Gauss, simply models dataset-level statistics, it is incapable of capturing scene-dependent error patterns.
Inspired by the \acp{pem} proposed in~\cite{Sadeghi2021NEURIPSW,Sadeghi2023ARXIV} we also design six object-conditioned baseline configurations that utilize a simple per-object feed-forward network to map the ground-truth state as well as a categorical visibility level~\cite{Caesar2019ARXIV} to a noisy detection output. Furthermore, we also construct a variant that additionally utilizes the same map encoder as our approach by concatenating the flattened map embedding to the per-object state projections.
Note that these methods fail to explicitly model false positives~\cite{Sadeghi2021NEURIPSW,Sadeghi2023ARXIV} and cannot capture error patterns that depend on other scene elements, such as duplicate detections. Unlike Static Gauss, they can, however, capture correlations between the ground-truth object state and detection errors on a per-object basis.
These baselines output the confidence scores for each class, as well as the parameters of a probability distribution for all regression targets. We train the models by optimizing the negative log likelihood for a given target detector. We use the same \ac{mlp} as \methodName{} for the input state projection, followed by either a three layer \ac{mlp} utilizing Layer normalization~\cite{ba2016layer} and an ELU activation function~\cite{clevert2015fast} or a ResNet~\cite{he2016deep} as proposed in~\cite{Sadeghi2023ARXIV} without dropout.
For the probability distribution, we either use a multivariate Gaussian as in~\cite{Sadeghi2021NEURIPSW} with a diagonal covariance matrix or a multivariate Student-T distribution as in~\cite{Sadeghi2023ARXIV} as the target distribution.  

\boldparagraph{Implementation Details}
For \methodName{}, we use $N_q^{gt} = 300$ queries for ground-truth, as well as $N_q^{ft} = 128$ queries for false positives. Each query has a latent dimensionality of $d_h = 256$ while the sampled latent code utilizes $d_z = 32$. This configuration is comparatively lightweight and achieves an inference throughput of roughly 190 scenes per second at batch size 64 on an Nvidia Titan Xp GPU. We choose $\beta = 0.01$ for weighting $\mathcal{L}_{JS}$ after a warm-up period of three epochs, while setting $\alpha = 0.5$ for an equal weighting of forward and backward KL.
We filter detector targets and \ac{pem} outputs for a minimum confidence score $\max(\bb^{cls}) \geq 0.2$ to ignore low confidence predictions. 

\boldparagraph{Results}
The performance of our proposed \ac{pem} for different detectors is shown in~\tabref{tab:pem_results}. Compared to both Static Gauss and the object-conditioned baselines, our model more precisely captures the error characteristics of the target detector, leading to consistent improvements in all metrics.
For DETR3D~\cite{Wang2021CORL}, which only uses inputs from a single time step, the CD-mPrec. is improved by $\SI{26}{\percent}$ over the best baseline.
Compared to target detectors that utilize temporal information, our proposed \ac{pem} improves the CD-mPrec. for BEVFormer~\cite{Li2022ECCV} as target by $\SI{32}{\percent}$ and by $\SI{15}{\percent}$ for StreamPETR~\cite{Wang2023ICCV} respectively. We also show the mean error characteristics for DETR3D in~\figref{fig:spider_plots}. Especially the reproduction of the translation error, velocity error and average precision is significantly improved in our model compared to the MLP + Gauss baseline.
\begin{table*}
\centering
\caption{\textbf{Effectiveness of Adversarial Perception Errors.} Left-hand side: comparison of baseline open-loop planning performance. Right-hand side: attained increase in \ac{cr} for varying latent space constraints $\kappa$.}
\begin{tabular}{l|ccc|ccc|ccc}
    \toprule
    & \multicolumn{3}{c|}{\textbf{Detector}}& \multicolumn{3}{c|}{\textbf{\ac{pem}}} & \multicolumn{3}{c}{\textbf{Adversarial \ac{pem}}}\\
     & CR & ADE & FDE & CR & ADE & FDE & \multicolumn{3}{c}{CR (\%)} \\
    \textbf{Model} & (\%) & (m) & (m) & (\%) & (m) & (m) & $\kappa = \num{1}$ & $\kappa = \num{2}$ & $\kappa = \num{3}$ \\
    \midrule
    Detr3d~\cite{Wang2021CORL} & 3.40 & 1.23 & 2.59 & 3.56 & 1.22 & 2.57 & 4.27\scriptsize{~(+20\%)} & 5.09\scriptsize{~(+43\%)} & 5.88\scriptsize{~(+65\%)}\\
    BEVFormer~\cite{Li2022ECCV} & 3.20 & 1.25 & 2.63 & 3.36 & 1.24 & 2.61 & 4.47\scriptsize{~(+33\%)} & 5.27\scriptsize{~(+57\%)} & 6.20\scriptsize{~(+85\%)}\\
    StreamPETR~\cite{Wang2023ICCV} & 3.40 & 1.27 & 2.67 & 3.58 & 1.25 & 2.60 & 4.77\scriptsize{~(+33\%)} & 5.35\scriptsize{~(+49\%)} & 6.28\scriptsize{~(+75\%)}\\
\bottomrule
\end{tabular}
\label{tab:attack_results}
\end{table*}

In~\figref{fig:quali_pem}, we show two qualitative examples of sampled perception errors for the MLP + Gauss baseline and \methodName{} in comparison to DETR3D~\cite{Wang2021CORL} as target detector.
In the first scene, the baseline model predicts implausible velocity estimates for oncoming traffic, while our approach models this correctly and samples realistic translation errors along line of sight as well reproducing similar orientation errors for large vehicles. The second scene highlights the importance of scene context. In contrast to our model, the baseline fails to sample plausible errors for the orientations of grouped parking vehicles, whilst our approach closely mimics the error modes of the target detector. Additional examples can be found in the supplementary material.

\boldparagraph{Ablation Study}
To verify our key design choices, the \emph{query initialization scheme} and \emph{choice of divergence}, we run experiments varying the number of false positive queries $N_q^{fp}$ and using the standard KL divergence instead of the skew-geometric JS divergence and report the results in~\tabref{tab:pem_results}. Doubling the number of false-positive queries results in similar performance at a higher computational burden, while omitting them drastically degrades performance, highlighting their importance. When using the KL- instead of the $\text{JS}^{G_\alpha}$ divergence we also see degraded performance. 
Additionally, we observe that it can permit high latent variance, providing an avenue for implausible attacks.
~
\begin{figure}
    \centering
    \subfloat[Not keeping safe distance]{\includegraphics[width=0.49\textwidth]{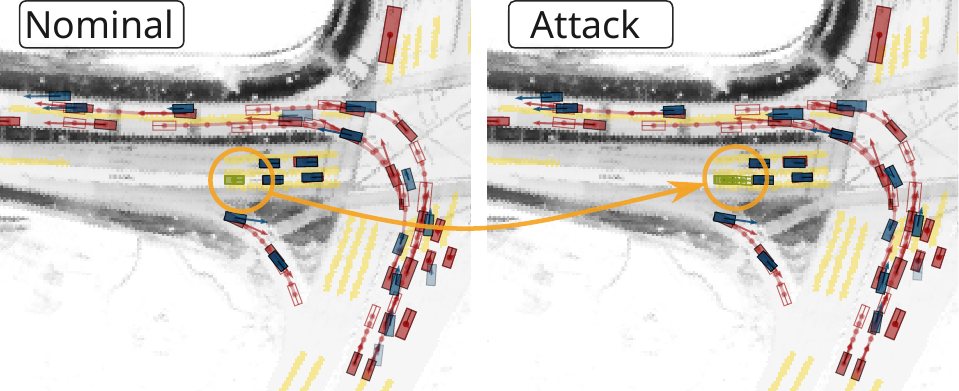}\label{fig:quali_attack:start}}
    \hfill%
    \subfloat[Dangerous sudden braking]{\includegraphics[width=0.49\textwidth]{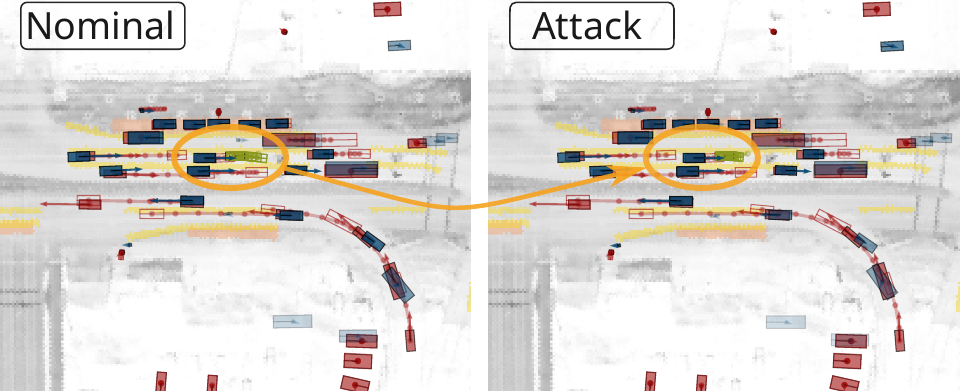}\label{fig:quali_attack:brake}}\\
    \caption{\textbf{Qualitative Examples of Worst-Case Perception Errors ($\kappa=3$).} Sampling adversarial perception errors from \methodName{} can induce unwarranted acceleration (\figref{fig:quali_attack:start}) and sudden braking (\figref{fig:quali_attack:brake}) in the planner, causing collisions. \textcolor{gtred}{Red} indicates ground truth objects, \textcolor{predblue}{blue} the model predictions. Non-filled boxes represent future states.}
    \label{fig:quali_attack}
\end{figure}

\subsection{Adversarial Perception Errors}
We now apply \methodName{} to probe the robustness of the \ac{il}-based planner described \secref{subsec:planner}. 
To this end, we run our proposed adversarial optimization procedure on scenes from the validation split, which is held-out for all involved models during training. 
To gain an understanding of its baseline performance, we first apply the planner to perception results obtained directly from the target detector. 
We also verify whether it performs similarly when operating on maximum likelihood samples obtained from \methodName{} (i.e. $\bz_n = \boldsymbol{\mu}^p_{\theta, n} (\bS, \cM) \ \forall \ \bz_n \in \bZ$). 
Finally, we optimize for failure over varying latent space constraints $\kappa$.

\boldparagraph{Metrics}
As we are interested in inducing unsafe behavior, we use the \acf{cr} as the main metric for this experiment. 
It measures the percentage of scenes for which the planned trajectory collides with another object within the planning horizon $T_\text{plan}$. 
This is supplemented by the \ac{ade}, which measures the average $l_2$-distance to the human expert trajectory within $T_\text{plan}$, as well as the \ac{fde}, which is similar to the \ac{ade} but considers only the final waypoint at $t=T_\text{plan}$.

\begin{figure}
    \centering
    \includegraphics[width=\columnwidth, trim={0.2cm 0.4cm 0.2cm 0.2cm},clip]{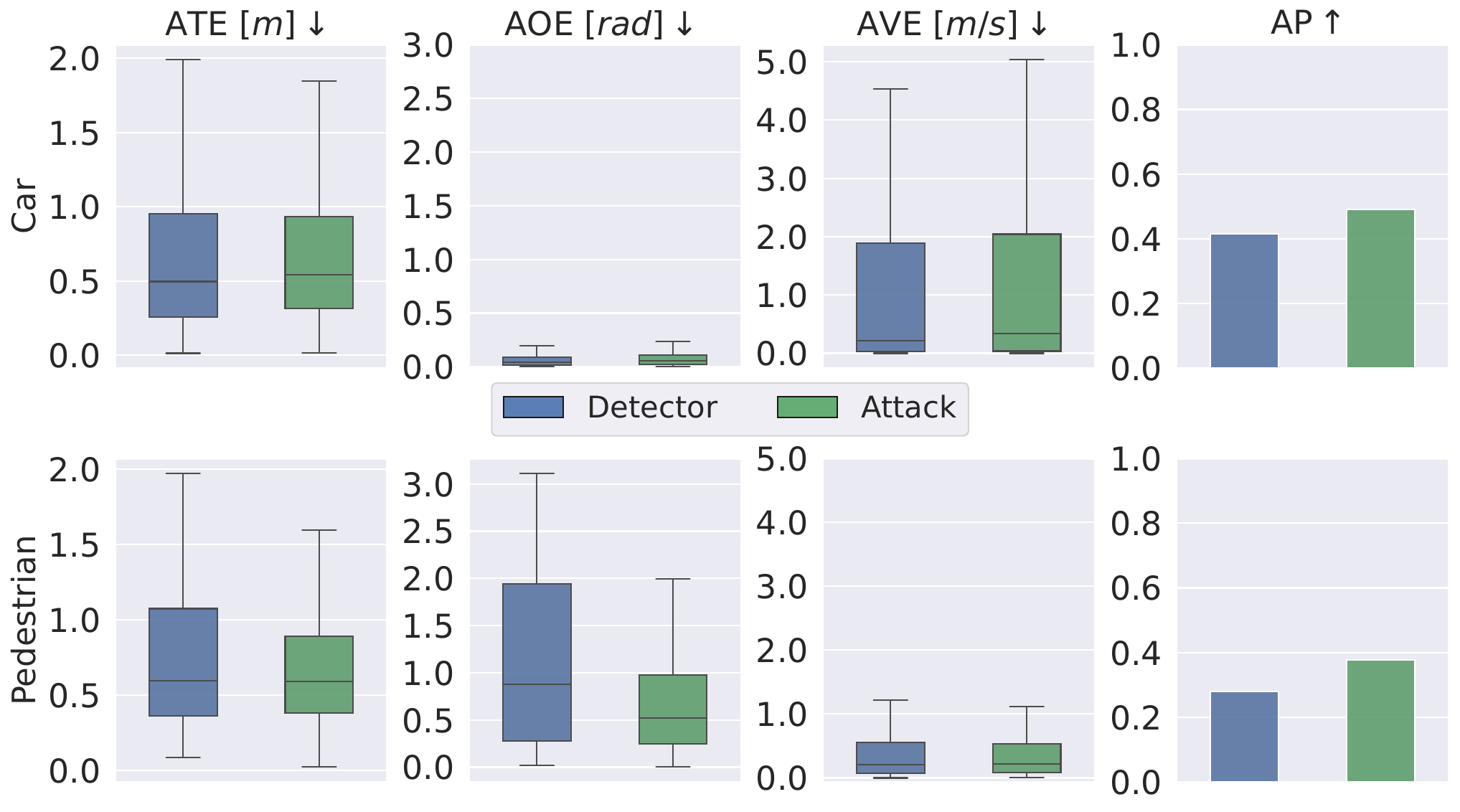}
    \caption{\textbf{Attack Plausibility.} Successful attacks (with $\kappa = 3$) show a similar or more accurate distribution of errors compared to DETR3D, highlighting the realism and conservative nature of our adversarial framework.
    }
    \label{fig:box_plots}
    \vspace{-0.2cm}
\end{figure}

\boldparagraph{Implementation Details}
We use the Adam~\cite{Kingma2015ICLR} optimizer with a learning rate of \num{1e-1} for the adversarial optimization procedure, which we run for a maximum of $N_\text{opt}=100$ iterations per scene, permitting $N_\text{trial}=5$ trials each. 
The strength of the likelihood regularization is set to $\lambda = \num{1e-1}$. We consider a planning horizon $T_\text{plan}$ of \SI{3}{\second} and train dedicated versions of the planner on inference results of each target detector using the tracking sub-split.

\boldparagraph{Results}
In addition to the evaluation presented in \secref{subsec:pem_eval}, the quantitative results reported in \tabref{tab:attack_results} further indicate \methodName{} to be a capable \ac{pem} in terms of downstream planning. 
Compared to the actual inference results of the target detector, operating the planner on maximum likelihood proxy samples results in similar a \ac{cr}, \ac{ade} and \ac{fde}. 
Furthermore, when allowing the adversarial optimization procedure to adjust the set of latent variables~$\bZ$, the baseline \ac{cr} can be increased by up to \num{85}\%, suggesting a critical vulnerability of our prototypical planner to even plausible perturbations in its input. 
The adversarial optimization most strongly affects the planned trajectory longitudinally, for example by inducing sudden dangerous braking or acceleration while in stationary traffic. We hypothesize this is due to a bias of heading straight-ahead that is common in driving data.
This is visualized in~\figref{fig:quali_attack}. Additional examples can be found in the supplementary material.

\boldparagraph{Attack Plausibility}
Through likelihood regularization~\eqnref{eq:adv_cost}, search space constraints~\eqnref{eq:adv_opt} and use of the JS- instead of KL-Divergence~\eqnref{eq:divergence_reg}, we design our method to be conservative such that it generates plausible attacks rather than extreme perturbations.
This is evident from~\figref{fig:box_plots}, where we show boxplots of real-world detector errors and \methodName{} samples inducing successful attacks (with $\kappa=3$) on the same sensor logs: Even after adversarial optimization, \methodName{} produces results that either match the detector distribution or show lower errors, both in terms of the median and spread of the distribution, highlighting their plausibility.

%% file: sections/5_conclusion.tex
\section{Conclusion}
We presented \methodName{}, a novel generative \acf{pem} that mimics the outputs of a perception system given a ground-truth scene representation. We have demonstrated its utility as an evaluation tool by probing an \acf{il}-based planner in terms of its robustness to noise in its inputs. Furthermore, we showed that \methodName{} more accurately captures the target detector's error characteristics than \acp{pem} used in previous work. However, there are remaining limitations opening avenues for improvement in future work. Firstly, we manually designed a cost function to induce a specific failure mode (i.e. collision). To enable inducing a wider range of planning failures, learning a general driving cost function from data~\cite{Arora2021IRLSurvey} is an interesting direction. Secondly, extending \methodName{} to model latent features instead of explicit intermediate representations, such as 3D bounding boxes, would enable stress-testing modular end-to-end trainable architectures, which have recently gained popularity~\cite{Hu2023CVPR, Jiang2023ICCV}.

%% file: root.bbl
\begin{thebibliography}{10}
\providecommand{\url}[1]{#1}
\csname url@samestyle\endcsname
\providecommand{\newblock}{\relax}
\providecommand{\bibinfo}[2]{#2}
\providecommand{\BIBentrySTDinterwordspacing}{\spaceskip=0pt\relax}
\providecommand{\BIBentryALTinterwordstretchfactor}{4}
\providecommand{\BIBentryALTinterwordspacing}{\spaceskip=\fontdimen2\font plus
\BIBentryALTinterwordstretchfactor\fontdimen3\font minus \fontdimen4\font\relax}
\providecommand{\BIBforeignlanguage}[2]{{%
\expandafter\ifx\csname l@#1\endcsname\relax
\typeout{** WARNING: IEEEtran.bst: No hyphenation pattern has been}%
\typeout{** loaded for the language `#1'. Using the pattern for}%
\typeout{** the default language instead.}%
\else
\language=\csname l@#1\endcsname
\fi
#2}}
\providecommand{\BIBdecl}{\relax}
\BIBdecl

\bibitem{Janai2020}
J.~Janai, F.~Güney, A.~Behl, and A.~Geiger, \emph{Computer Vision for Autonomous Vehicles: Problems, Datasets and State of the Art}.\hskip 1em plus 0.5em minus 0.4em\relax Foundations and Trends in Computer Graphics and Vision, 2020, vol.~12, no. 1-3.

\bibitem{Wang2021CORL}
Y.~Wang, V.~. Guizilini, T.~Zhang, Y.~Wang, H.~Zhao, and J.~M. Solomon, ``Detr3d: 3d object detection from multi-view images via 3d-to-2d queries,'' in \emph{CoRL}, 2021.

\bibitem{Wang2023ICCV}
S.~Wang, Y.~Liu, T.~Wang, Y.~Li, and X.~Zhang, ``Exploring object-centric temporal modeling for efficient multi-view 3d object detection,'' in \emph{ICCV}, 2023.

\bibitem{Li2023Toponet}
T.~Li, L.~Chen, H.~Wang, Y.~Li, J.~Yang, X.~Geng, S.~Jiang, Y.~Wang, H.~Xu, C.~Xu, J.~Yan, P.~Luo, and H.~Li, ``Graph-based topology reasoning for driving scenes,'' \emph{arXiv.org}, vol. 2304.05277, 2023.

\bibitem{Salzmann2020ECCV}
T.~Salzmann, B.~Ivanovic, P.~Chakravarty, and M.~Pavone, ``Trajectron++: Dynamically-feasible trajectory forecasting with heterogeneous data,'' in \emph{ECCV}, 2020.

\bibitem{Hu2021FIERY}
A.~Hu, Z.~Murez, N.~Mohan, S.~Dudas, J.~Hawke, V.~Badrinarayanan, R.~Cipolla, and A.~Kendall, ``{FIERY}: Future instance segmentation in bird's-eye view from surround monocular cameras,'' in \emph{ICCV}, 2021.

\bibitem{Renz2022CORL}
K.~Renz, K.~Chitta, O.-B. Mercea, A.~S. Koepke, Z.~Akata, and A.~Geiger, ``Plant: Explainable planning transformers via object-level representations,'' in \emph{CoRL}, 2022.

\bibitem{Dauner2023Parting}
D.~Dauner, M.~Hallgarten, A.~Geiger, and K.~Chitta, ``Parting with misconceptions about learning-based vehicle motion planning,'' in \emph{CoRL}, 2023.

\bibitem{Caesar2021CVPRWORK}
H.~Caesar, J.~Kabzan, K.~S. Tan, W.~K. Fong, E.~Wolff, A.~Lang, L.~Fletcher, O.~Beijbom, and S.~Omari, ``Nuplan: A closed-loop ml-based planning benchmark for autonomous vehicles,'' in \emph{CVPR Workshops}, 2021.

\bibitem{Montali2023ARXIV}
N.~Montali, J.~Lambert, P.~Mougin, A.~Kuefler, N.~Rhinehart, M.~Li, C.~Gulino, T.~Emrich, Z.~Yang, S.~Whiteson, B.~White, and D.~Anguelov, ``The waymo open sim agents challenge,'' \emph{arXiv.org}, vol. 2305.12032, 2023.

\bibitem{Filos2020ICML}
A.~Filos, P.~Tigas, R.~McAllister, N.~Rhinehart, S.~Levine, and Y.~Gal, ``Can autonomous vehicles identify, recover from, and adapt to distribution shifts?'' in \emph{ICML}, 2020.

\bibitem{Innes2023ICRA}
C.~Innes and S.~Ramamoorthy, ``Testing rare downstream safety violations via upstream adaptive sampling of perception error models,'' in \emph{ICRA}, 2023.

\bibitem{Sadeghi2023ARXIV}
J.~Sadeghi, N.~A. Lord, J.~Redford, and R.~Mueller, ``Attacking motion planners using adversarial perception errors,'' \emph{arXiv.org}, vol. 2311.12722, 2023.

\bibitem{Piazzoni2021IJCAI}
A.~Piazzoni, J.~Cherian, M.~Slavik, and J.~Dauwels, ``Modeling perception errors towards robust decision making in autonomous vehicles,'' in \emph{IJCAI}, 2021.

\bibitem{Sadeghi2021NEURIPSW}
J.~Sadeghi, B.~Rogers, J.~Gunn, T.~Saunders, S.~Samangooei, P.~K. Dokania, and J.~Redford, ``A step towards efficient evaluation of complex perception tasks in simulation,'' in \emph{NeurIPS Workshops}, 2021.

\bibitem{Mitra2018ITSC}
P.~Mitra, A.~Choudhury, V.~R. Aparow, G.~Kulandaivelu, and J.~Dauwels, ``Towards modeling of perception errors in autonomous vehicles,'' in \emph{ITSC}, 2018.

\bibitem{Piazzoni2023ARXIV}
A.~Piazzoni, J.~Cherian, J.~Dauwels, and L.-P. Chau, ``On the simulation of perception errors in autonomous vehicles,'' \emph{arXiv.org}, vol. 2302.11919, 2023.

\bibitem{Wong2020ECCV}
K.~Wong, Q.~Zhang, M.~Liang, B.~Yang, R.~Liao, A.~Sadat, and R.~Urtasun, ``Testing the safety of self-driving vehicles by simulating perception and prediction,'' in \emph{ECCV}, 2020.

\bibitem{Ju2023PerceptionImitation}
X.~Ju, Y.~Sun, Y.~Hao, Y.~Li, Y.~Qiao, and H.~Li, ``Perception imitation: Towards synthesis-free simulator for autonomous vehicles,'' \emph{arXiv.org}, vol. 2304.09365, 2023.

\bibitem{Yang2018PIXOR}
B.~Yang, W.~Luo, and R.~Urtasun, ``Pixor: Real-time 3d object detection from point clouds,'' in \emph{CVPR}, 2018.

\bibitem{Yin2021Centerpoint}
T.~Yin, X.~Zhou, and P.~Kr{\"a}henb{\"u}hl, ``Center-based 3d object detection and tracking,'' \emph{CVPR}, 2021.

\bibitem{Ding2020IROS}
W.~Ding, M.~Xu, and D.~Zhao, ``Learning to collide: An adaptive safety-critical scenarios generating method,'' in \emph{IROS}, 2020.

\bibitem{Suo2023CVPR}
S.~Suo, K.~Wong, J.~Xu, J.~Tu, A.~Cui, S.~Casas, and R.~Urtasun, ``Mixsim: A hierarchical framework for mixed reality traffic simulation,'' in \emph{CVPR}, 2023.

\bibitem{Wang2021CVPRb}
J.~Wang, A.~Pun, J.~Tu, S.~Manivasagam, A.~Sadat, S.~Casas, M.~Ren, and R.~Urtasun, ``Advsim: Generating safety-critical scenarios for self-driving vehicles,'' in \emph{CVPR}, 2021.

\bibitem{Rempe2022CVPR}
D.~Rempe, J.~Philion, L.~J. Guibas, S.~Fidler, and O.~Litany, ``Generating useful accident-prone driving scenarios via a learned traffic prior,'' in \emph{CVPR}, 2022.

\bibitem{Hanselmann2022ECCV}
N.~Hanselmann, K.~Renz, K.~Chitta, A.~Bhattacharyya, and A.~Geiger, ``{KING}: Generating safety-critical driving scenarios for robust imitation via kinematics gradients,'' in \emph{ECCV}, 2022.

\bibitem{Carion2020ECCV}
N.~Carion, F.~Massa, G.~Synnaeve, N.~Usunier, A.~Kirillov, and S.~Zagoruyko, ``End-to-end object detection with transformers,'' in \emph{ECCV}, 2020.

\bibitem{Sohn2015NIPS}
K.~Sohn, H.~Lee, and X.~Yan, ``Learning structured output representation using deep conditional generative models,'' \emph{NIPS}, 2015.

\bibitem{Kingma2014ICLR}
D.~P. Kingma and M.~Welling, ``Auto-encoding variational bayes,'' \emph{ICLR}, 2014.

\bibitem{Li2022ECCV}
Z.~Li, W.~Wang, H.~Li, E.~Xie, C.~Sima, T.~Lu, Y.~Qiao, and J.~Dai, ``Bevformer: Learning bird’s-eye-view representation from multi-camera images via spatiotemporal transformers,'' in \emph{ECCV}, 2022.

\bibitem{Suo2021CVPR}
S.~Suo, S.~Regalado, S.~Casas, and R.~Urtasun, ``Trafficsim: Learning to simulate realistic multi-agent behaviors,'' in \emph{CVPR}, 2021.

\bibitem{Caesar2019ARXIV}
H.~Caesar, V.~Bankiti, A.~H. Lang, S.~Vora, V.~E. Liong, Q.~Xu, A.~Krishnan, Y.~Pan, G.~Baldan, and O.~Beijbom, ``nuscenes: {A} multimodal dataset for autonomous driving,'' \emph{arXiv.org}, 2019.

\bibitem{Deasy2020NEURIPS}
J.~Deasy, N.~Simidjievski, and P.~Li{\`o}, ``Constraining variational inference with geometric jensen-shannon divergence,'' in \emph{NeurIPS}, 2020.

\bibitem{Higgins2017ICLR}
I.~Higgins, L.~Matthey, A.~Pal, C.~Burgess, X.~Glorot, M.~Botvinick, S.~Mohamed, and A.~Lerchner, ``beta-vae: Learning basic visual concepts with a constrained variational framework,'' in \emph{ICLR}, 2017.

\bibitem{Hu2023CVPR}
Y.~Hu, J.~Yang, L.~Chen, K.~Li, C.~Sima, X.~Zhu, S.~Chai, S.~Du, T.~Lin, W.~Wang, L.~Lu, X.~Jia, Q.~Liu, J.~Dai, Y.~Qiao, and H.~Li, ``Planning-oriented autonomous driving,'' in \emph{CVPR}, 2023.

\bibitem{Jiang2023ICCV}
B.~Jiang, S.~Chen, Q.~Xu, B.~Liao, J.~Chen, H.~Zhou, Q.~Zhang, W.~Liu, C.~Huang, and X.~Wang, ``Vad: Vectorized scene representation for efficient autonomous driving,'' \emph{ICCV}, 2023.

\bibitem{Doll2024DualAD}
S.~Doll, N.~Hanselmann, L.~Schneider, R.~Schulz, M.~Cordts, M.~Enzweiler, and H.~P. Lensch, ``Dualad: Disentangling the dynamic and static world for end-to-end driving,'' in \emph{CVPR}, 2024.

\bibitem{he2016deep}
K.~He, X.~Zhang, S.~Ren, and J.~Sun, ``Deep residual learning for image recognition,'' in \emph{Proceedings of the IEEE conference on computer vision and pattern recognition}, 2016, pp. 770--778.

\bibitem{ba2016layer}
J.~L. Ba, J.~R. Kiros, and G.~E. Hinton, ``Layer normalization,'' \emph{arXiv.org}, vol. 1607.06450, 2016.

\bibitem{clevert2015fast}
D.-A. Clevert, T.~Unterthiner, and S.~Hochreiter, ``Fast and accurate deep network learning by exponential linear units (elus),'' \emph{arXiv.org}, vol. 1511.07289, 2015.

\bibitem{Kingma2015ICLR}
D.~P. Kingma and J.~Ba, ``Adam: {A} method for stochastic optimization,'' in \emph{ICLR}, 2015.

\bibitem{Arora2021IRLSurvey}
S.~Arora and P.~Doshi, ``A survey of inverse reinforcement learning: Challenges, methods and progress,'' \emph{AI}, vol. 297, p. 103500, 2021.

\end{thebibliography}
